\documentclass{article}
\usepackage{fnfm}

\usepackage{times}
\usepackage{xcolor}
\usepackage{soul}
\usepackage[small]{caption}
\usepackage{subfigure}
\usepackage{graphicx}
\usepackage{amsthm}
\usepackage{epsfig,algorithmic,algorithm}
\usepackage{graphicx}
\usepackage{subfigure}
\usepackage{multirow}

\usepackage{amsmath,bm}
\usepackage{amssymb}

\usepackage[english]{babel}

\usepackage{arydshln}

\title{Field-aware Neural Factorization Machine for Click-Through Rate Prediction}

\author{Li Zhang$^{1}$, Weichen Shen$^{1}$, Shijian Li$^{1*}$, Gang Pan$^{1}$\\
$^1$Zhejiang University, Hangzhou, China\\
\{zhangli85,weichenshen,shijiangli,gpan\}@zju.edu.cn}
\begin{document}

\maketitle

\begin{abstract}
Recommendation systems and computing advertisements have gradually entered the field of academic research from the field of commercial applications. Click-through rate prediction is one of the core research issues because the prediction accuracy affects the user experience and the revenue of merchants and platforms. Feature engineering is very important to improve click-through rate prediction. Traditional feature engineering heavily relies on people’s experience, and is difficult to construct a feature combination that can describe the complex patterns implied in the data. This paper combines traditional feature combination methods and deep neural networks to automate feature combinations to improve the accuracy of click-through rate prediction. We propose a mechannism named 'Field-aware Neural Factorization Machine' (FNFM). This model can have strong second order feature interactive learning ability like Field-aware Factorization Machine, on this basis, deep neural network is used for higher-order feature combination learning. Experiments show that the model has stronger expression ability than current deep learning feature combination models like the DeepFM, DCN and NFM.

\end{abstract}

\section{Introduction}

The recommendation system was developed to address the demands of users and businesses in the Internet scene. According to the data, the recommendation system brought 35\% of sales revenue to Amazon, and 75\% of Netflix's consumption. 60\% of the browsing traffic on the Youtube homepage comes from personalized recommendation traffic. Therefore, building accurate and effective recommendation systems is of great significance for improving user experience and company revenue.

In the recommendation system, a crucial task is to predict the probability of a user clicking on a recommended item. Therefore, click-through rate prediction is a core issue for recommendation systems. In many recommendation systems, the goal is to maximize the number of clicks, so recommended items can be ranked by estimated clickthrough rate. In addition, in online advertising systems \cite{pepelyshev2015adaptive}, click-through rate prediction is also very important to improve system revenue, because the ad's sorting strategy can be adjusted by clickthrough rate and bidding.

Features play a central role in the success of many predictive systems. Different features present specific information from various aspects and dimensions, and cross-combination between features is often very meaningful. Traditional cross-over features have three major drawbacks: First, obtaining high quality features require high cost. Since effective feature combinations are often generated based on specific task scenarios, engineers need to spend a lot of time manually designing cross-combination features, and artificial feature engineering relies heavily on engineers' prior knowledge and business sensitivity, which has great limitations. Second, in large-scale prediction systems such as recommendation systems, a large number of original features make manual extraction of all cross-features infeasible. Finally, the artificially constructed cross-combination features cannot be overlaid onto the combined patterns in the training data that have occurred. The deep learning technology is a promising way to solve the problems, since it has advantage to handle inner sturctures inside high-dimensional sparse data scenarios\cite{broder2008computational}, using deep learning technology to improve the feature interaction ability of predictive models is a meaningful research task.

Several works was done for automatic feature engineering with deep neural networks, like the Deep Factorization Machine (DeepFM), Neural Factorization Machine and Deep Cross Networks. Inspired by these works, we considered more about further reduction of information loss and confusion in feature combination, and proposed our factorization model named 'Field-aware Neural Factorization Machine' (FNFM). FNFM embeds information in units of 'fields' and use field vector to present the input information, so it can retain more information in second and higher order feature interactive learning.Experiments show FNFM has better information expressive ability compared with other models.  

\section{Related Work}

Traditional linear models capture second-order feature combinations by means of degree-2 polynomial features. Data sparsity widely exists in actual business scenarios. For some combination features, there are very few cases where the two features are not zero at the same time. When any feature in the combined feature takes a value of 0, then other features. The interaction term coefficients of the combination with this feature cannot be effectively learned.

The factorization machine\cite{rendle2010factorization} uses the idea of ​​matrix decomposition \cite{lee2013local} to obtain the matrix of interaction term coefficients between features by the implicit inner product of the feature. This method makes up for the shortcomings of the second-order polynomial method that the amount of parameters that need to be learned is too large and cannot effectively process the coefficient data.

The FFM(Field-aware Factorization Machine) \cite{juan2016field} model is an improvement of the FM model. The FFM model introduces the concept of field, that is, using different hidden vectors presenting different feature groups. When calculating the weight of the interaction term between each pair of features, the traditional FM model is represented by the inner product of the hidden vectors corresponding to the two features. 

In the FFM, each feature $x_i$ in feature group $i$, for features $x_j$ in other feature group $j$, FFM learns a pair of hidden vectors $v_{i,f_j }, v_{j,f_i}$. It divides it into multiple fields according to the meaning of the feature, and each feature belongs to a specific field. Each feature has multiple hidden vectors, one for each field. When two features are combined, the inner product of the field corresponding to the two features is used as the inner product, so the model equation of FFM is:

\begin{equation}
    \hat{y}=w_0+\sum_{i=1}^nw_ix_i+\sum_{i=1}^n\sum_{j=i+1}^n<v_{i,f_j},v_{j,f_i}>x_ix_j
\end{equation}

Where $f_i$ and $f_j$ are the fields to which the $i$th and $j$th features belong, respectively. If there are a total of $f$ fields, then the parameter quantity of the FFM model is $nfk$, and the computation time complexity is $O(\bar{n}^2k$). It is worth noting that in FFM, each hidden vector only needs to learn the effect of interacting with a specific field vector, 

DeepFM\cite{guo2017deepfm} is a model that combines FM and DNN to model low-order feature combinations like FM and model high-order feature combinations like DNN. Unlike WDL\cite{cheng2016wide}, DeepFM can perform end-to-end training without any feature engineering because its wide side and deep side share the same input and embedding vectors. The model structure is as follows:

\begin{figure}[H]
\begin{center}
\includegraphics[scale=0.27]{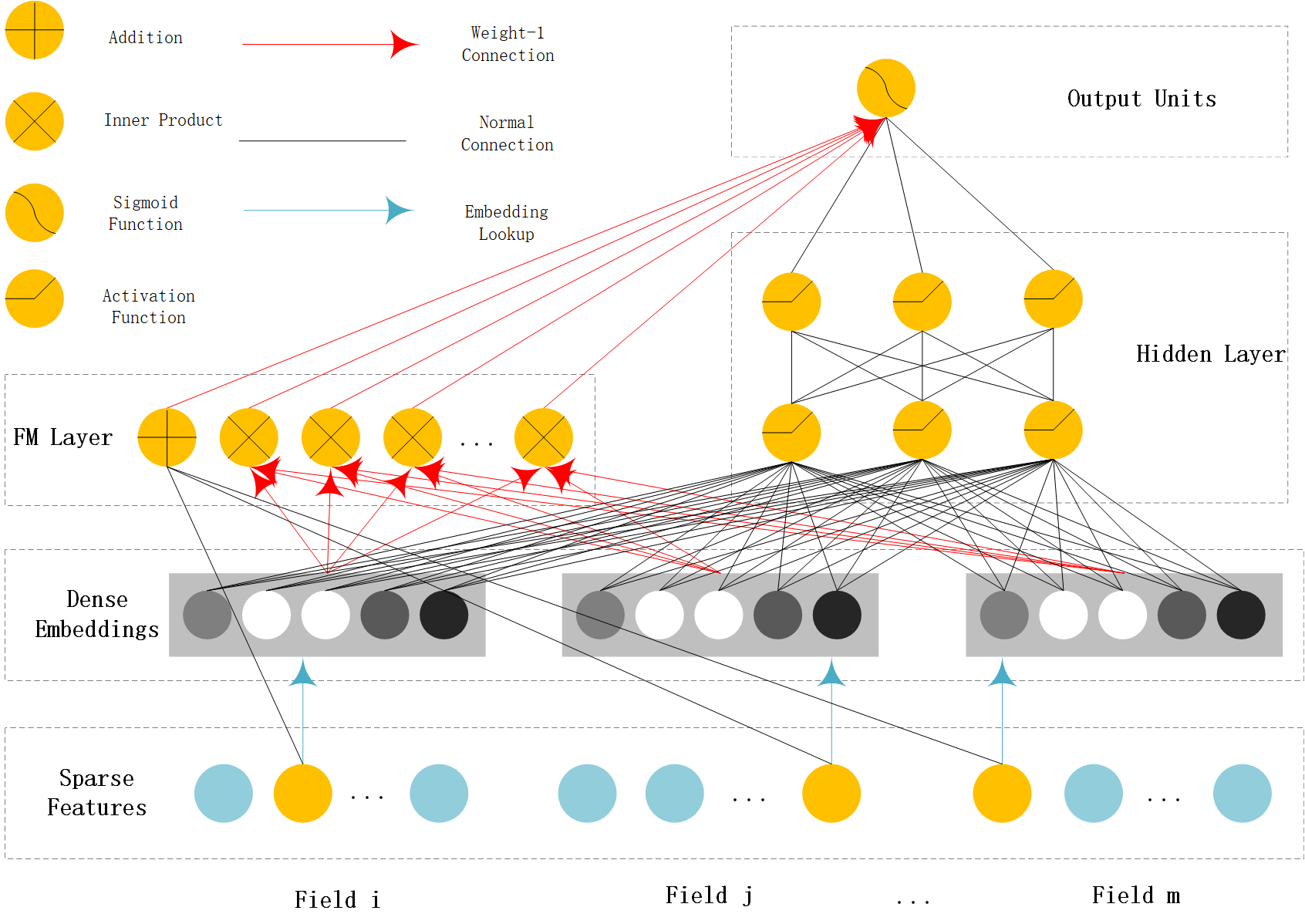}
\end{center}
\vspace*{-15pt}
   \caption{DeepFM Model Sturucure}
\label{fig:Model Structure}
\vspace*{-12pt}
\end{figure}

DeepFM consists of two components that share the same input FM component and DNN component. For feature $x_i$, a scalar $w_i$ is used as its 1-order weight, and a hidden vector $V_i$ is used as an influence factor for its interaction with other features. $V_i$ is input into the FM component to model the 2-order feature interaction, while inputting into the DNN component to model higher-order feature interactions. All parameters are trained by joint prediction models:
\begin{equation}
    \hat{y} = sigmoid(y_{FM}+y_{DNN})
\end{equation}

The NFM(Neural FM)\cite{he2017neural} model uses both FM and neural networks to model sparse data.

\begin{figure}[H]
\begin{center}
\includegraphics[scale=0.27]{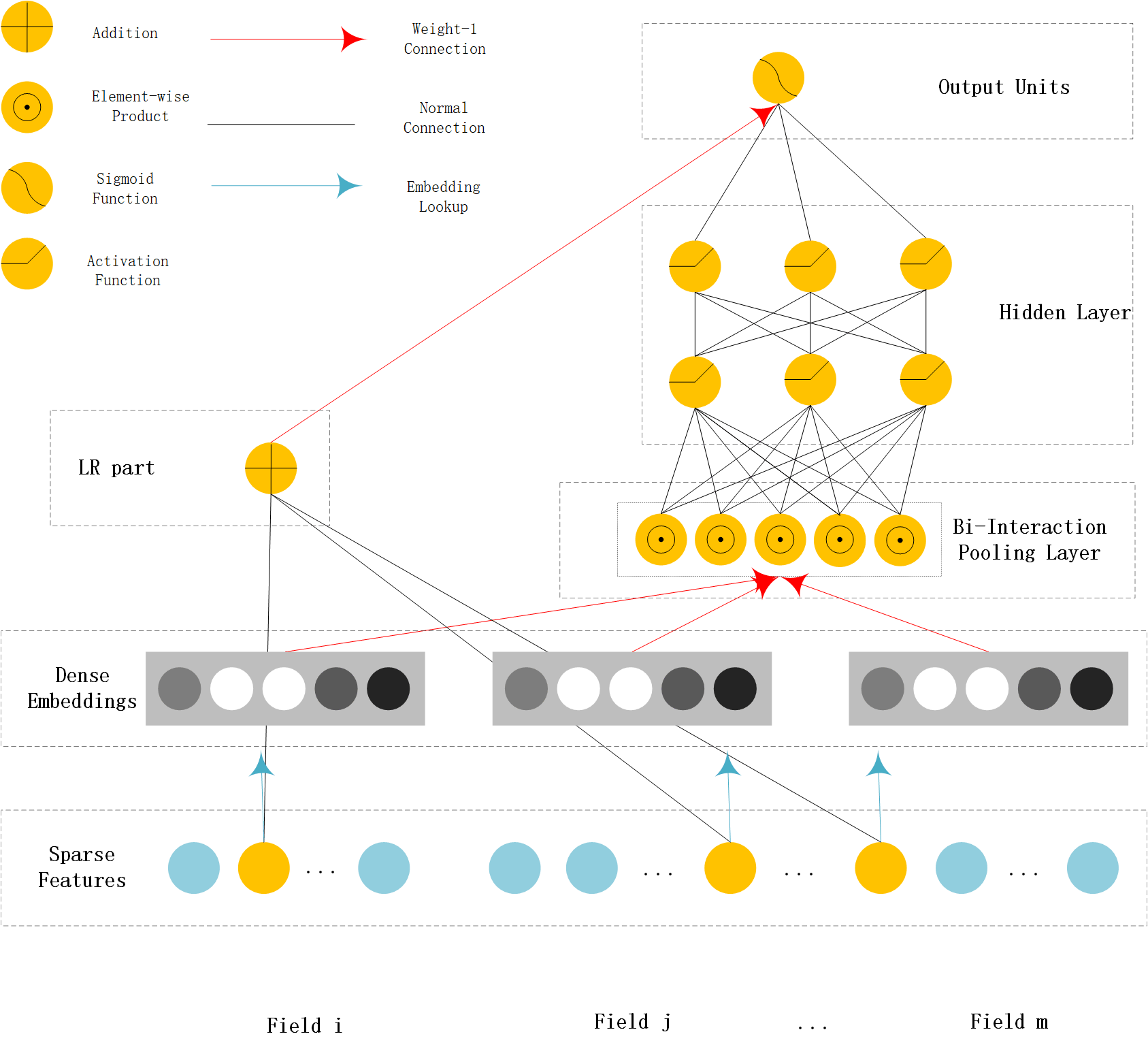}
\end{center}
\vspace*{-15pt}
   \caption{NFM Model Sturucure}
\label{fig:Model Structure}
\vspace*{-12pt}
\end{figure}

\begin{equation}
    \hat{y} = w_0 + \sum_{i=1}^n{w_ix_i}+f_{BI}(x)
\end{equation}

In the equation, the first and second items are similar to the linear regression items in the FM model. The third item$f_{BI}(x)$ is the core component used by the NFM model to model feature interactions.
NFM first incorporates it into the vector input to a second-order interactive pooling layer, which is capable of pooling several embedded vectors into a vector:

\begin{equation}
    f_{BI}(V_x) =\sum_{i=1}^n\sum_{j=i+1}^n{x_iv_i}\odot{x_jv_j}
\end{equation}

Where $\odot$ represents the element-wise multiplication between two vectors. The output of the second-order interaction layer is a $k$-dimensional vector that encodes the second-order interaction between features into the embedded space.

Deep and Cross Network (DCN)\cite{wang2017adkdd} adopts a cross-network structure to explicitly calculate the cross-combination between features. The crossover network consists of different intersecting layers. This special structure enables the order of the interactive features to increase as the number of layers increases. The highest order (as compared to the original input) that a $L$ layered cross network can capture is $L+1$..

Deep networks get factorition machines the ability of higher ordered interaction, but because the interaction between features are based on feature elements which may confuse the information betten features. We consider to combine filed-based FFM and deep neural network to generate an expressive factorization model.

\section{Field-aware Neural Factorization Machine}

We propose Field-aware Neural Factorization Machine (FNFM), a click-through prediction model which obtains the advantages of FFM in second order feature interaction and improves NFM's ability on higher order feature interaction. The model of FNFM is described as:
\begin{equation}
\widehat{y}_{FNFM}(x)=w_{o}+\sum_1^nw_ix_i+DNN(f_{BI}(V_x))
\end{equation}

The first two items is same regression function as FFM, and the last item in equation $DNN(f_{BI}(V_x))$ is the key mechanism to handle feature interactions. The mechanism of FNFM consists of four layers. Lower parts works like FFM, and higher is DNN, comparing with DFM, our model has information concatation and mormalization layers.

\begin{figure}[H]
\begin{center}
\includegraphics[scale=0.27]{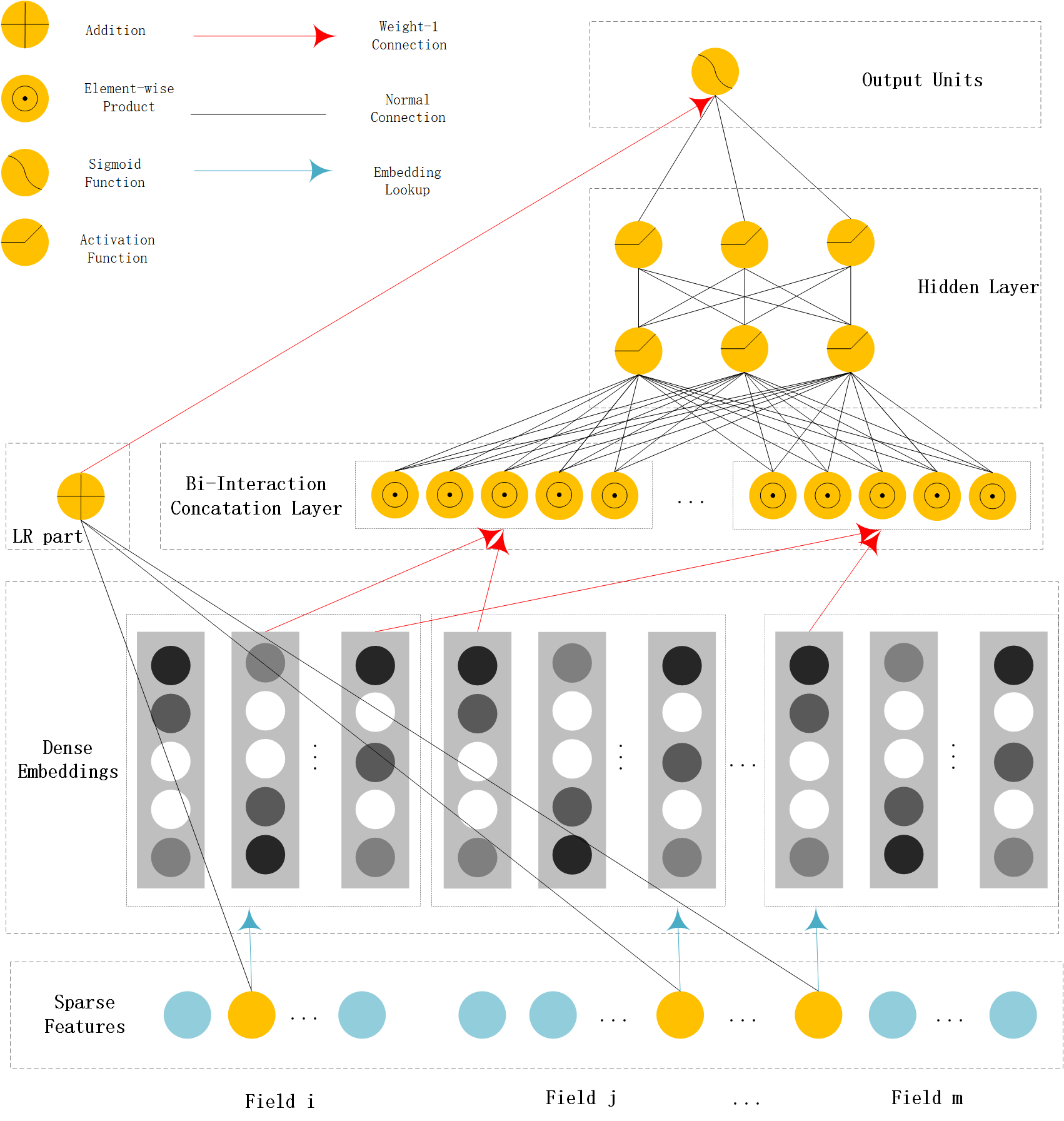}
\end{center}
\vspace*{-15pt}
   \caption{FNFM Model Sturucure}
\label{fig:Model Structure}
\vspace*{-12pt}
\end{figure}

\subsection{Input Layer}

We first convert the characteristics of users and advertisements into a feature vector that is spliced ​​from input features under different feature groups (as fields in FFM):
\begin{equation}
    x = [x_1;x_2;\dots;x_f] 
\end{equation}

Where $f$ is the number of feature groups.$x_i$ is the feature of  $t$-th feature group.If it is a sparse category feature, then $x_i$ is a one hot encoding vector. If the $t$th feature group is a dense numerical feature feature group, then the $x_i$ is a scalar.

\subsection{Embedding Layer}

Since the feature representations of category features are generally high-dimensional and sparse, they are usually
It is compressed into a low dimensional space. Traditional embedding techniques map feature vectors under each feature set to:

\begin{equation}
    e_i = V_ix_i 
\end{equation}

Where $V_i$ is the embedding matrix corresponding to feature group $t_i$, and $x_i$ is a onehot encoding vector.
In order to be able to cross and combine dense numerical features with sparse category features, dense numerical features can also be compressed into low-dimensional spaces by embedding techniques. Express numerical features as:

\begin{equation}
    e_m = v_mx_m
\end{equation}

Where $v_m$ is an embedding vector corresponding to a feature set m, and $x_m$ is a scalar input value. By projecting the category and numerical features together into a low-dimensional space of the same dimension, it can rlearn the interaction between features under different feature groups through cross layer.

The embedded layer of the FNFM model is imported using a feature group embedding method similar to the FFM model.
Convert to a low-dimensional dense representation, for the $i$-th feature group $t_i$, let $V^{ij} = [v_1^{ij},...,v_m^{ij},...,v_{K_i}^{ij}] \in R^{D\times K_i}$ 
represent the embedded vector dictionary used by the $i$-th feature group to interact with features under the $j$-th feature gourp, where $v_m^{ij} \in R^D$ is a $D$ dimension embedded vector.

\subsection{Bi-Interaction Concatation Layer}

The FNFM model uses the idea of ​​factorization to learn the expression of second-order interactive features in the form of hidden vector products. Different from the traditional models such as DeepFM and NFM, the FNFM model uses a second-order feature interaction method based on field aware method.

Let two input features from different feature groups $f_i$, $f_j$ be divided into $x_i, x_j$, and the second-order feature interaction vector calculated by FNFM model is

\begin{equation}
a_{i,j}=x_iv_{i,f_j}\odot x_jv_{j,f_i}
\end{equation}

Where $\odot$denotes a vector element-by-element product operation, $v_{i,f_j}$ denotes an implicit vector used when the input $x_i$ interacts with input from the $f_j$th feature group, and $v_{j,f_i}$ denotes that the input $x_j$ interacts with the input from the $f_i$th feature group.

For a model with $f$ feature group inputs, we can find two second-order cross-product vectors of $\frac{f*(f-1)}{2}$, and use the pooling method to compress $\frac{f*(f-1)}{2}$ vectors into one with the NFM model. The $D$ dimension vector is different as the deep neural network input. The FNFM model uses a vector concatenate method to concat them into a vector of $\frac{f*(f-1)}{2}*D$ dimension:
\begin{equation}
    f_{BI}(V_x) = a_{1,2}  \oplus a_{1,3} \oplus \dots \oplus a_{f-1,f}
\end{equation}

Where $a_{i,j}$ are the intersection vectors of the features of the feature group $t_i$ and the feature group $t_j$, and $\oplus$ is the concatenate operator.
Compared with the traditional second-order interactive vector pooling layer, the second-order interactive concatenate layer builds a pooling for each second-order interaction , so it can retain the maximum information.
The information contained in the second-order interaction vector is beneficial to the subsequent deep neural network to extract higher-order combination modes.

\subsection{Normalization Layer}
Due to the use of field-aware embedded layer and interactive vector concatenate operations, the numerical statistical distribution of each interaction vector is insensitive to other interaction vectors in the model learning process.The numerical distribution of the output vectors of the second-order interactive concatenate layer in each dimension will have a large difference. 

This will reduce the overall convergence speed and performance of the model. We use batch nomalization in the second-order interactive concatenate layer to ensure that the statistical distribution of each dimension of the input vector of the deep neural network has small differences. Batch nomalization \cite{ioffe2015batch} is an adaptive reparameterization method that is mainly used to solve the problem that the gradient of the deep neural network disappears during the training process. Batch nomalization can transform the data into a statistical distribution with a mean of 0 variance of 1. If the input data of the network changes too much,using BN can make the input of the network more stable, because it make the learning of one layer independent with other parts of the network.

\subsection{Multiple Layer Perceptron (MLP)}
The MLP is used to extract high-ordered features and for prediction. Its input is concatenated concatenate embedding after batch normalization. Each layers in the MLP with activation function of RELU on last layer's output, and it uses softmax layer at last to complete the task of probability prediction.
\subsection{Loss Function}
The negative log-likelihood function is widely used in CTR models, which is usually defined as:

\begin{equation} \label{eq:loss function}
\begin{split}
L = - \frac{1}{N} \sum_{(x, y) \in \mathbb{S}} (y\,log p(x) + (1 - y)\,log(1 - p(x)))
\end{split}
\end{equation}

where $S$ is training data set whose size is $N$, $x$ is the input of the network, $y \in { 0, 1 }$ represents whether user clicks the item and $p(x)$ is the final output of the network which represents the probability that user will click the item.

\section{Experiment}
In this section, we move forward to evaluate the effectiveness of the performance of FNFM.

\subsection{Task and Data }

We uses Kaggle Avazu to display the ad click rate prediction data set, which contains a total of 40,428,967 samples in 10 days, including 33,563,901 positive samples and 68,865,66 negative samples. Due to the limited performance of the experimental machine, the experiment samples 10\% of the sample of the data set, selects the first 9 days as the training data set in the sampled samples, and divides 50\% of the samples into the verification data set and the test data set on the last day. The sampled data set has a total of 4042897 samples, including 3620824 training data sets, 211303 verification data sets, and 211037 test data sets.

\subsection{Bi-Interaction Concatation Layer VS Bi-Interaction Pooling Layer }

This section of the experiment will compare the performance of the FNFM model with different Bi-Interaction Layers.

All of the models in this section were targeted at minimizing the cross entropy loss function and optimized using the Adam method with a learning rate of 0.001. For the comprehensive consideration of training time and convergence speed, the batch size is chosen to be 4096. The network structure uses 3 hidden layers with 256 neurons per layer. Each feature group has a feature embedding dimension of 4 dimensions, an L2 regularization term with an intensity of 0.00001 for linear weights, and an L2 regularization term of 0.00001 for the embedded vector, and the hidden layer neurons do not use regularization terms. In addition, Batch Normalization technology is used on the output of the Bi-Interaction Layer.

The figure below shows the training error and verification error for each round of the FNFM model using different Bi-Interaction Layers.

\begin{figure}[H]
\begin{center}
\includegraphics[scale=0.6]{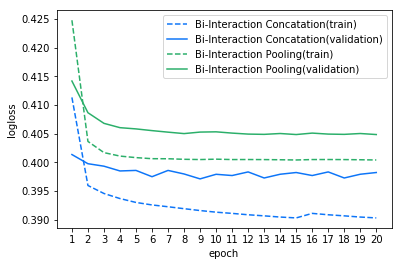}
\end{center}
\vspace*{-10pt}
   \caption{Comparison of learning curve between concatation layer and pooling layer
}
\label{fig:concatation layer}
\vspace*{-12pt}
\end{figure}

As you can see from the figure, the Bi-Interaction Layer using Concatation mode can achieve lower training and verification errors, which means that the Bi-Interaction Concatation Layer is used in the FNFM model compared to the Bi-Interaction Pooling Layer used in NFM can help improve the expressiveness of the model.

\subsection{ Batch Normalization }

We found that due to the use of the Bi-Concatation Layer, the distribution of input data to deep neural networks becomes unstable. The following figure shows the standard deviation distribution of neurons input by deep neural networks when using BN and not using BN:

\begin{figure}[H]
\begin{center}
\includegraphics[scale=0.6]{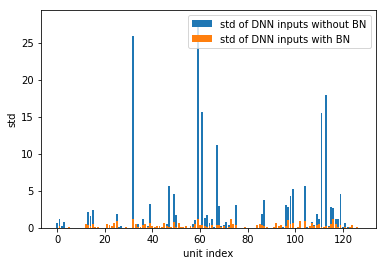}
\end{center}
\vspace*{-10pt}
   \caption{
DNN input standard deviation distribution when using BN and not using BN}
\label{fig:why normalization}
\vspace*{-12pt}
\end{figure}

As can be seen from the above figure, when BN is not used, the standard deviation of the input values ​​of the deep neural network is large, which indicates that the statistical distribution of the input data of the deep neural network is unstable. That will affect the learning process of subsequent network. And after using BN, the standard deviation of the input values is significantly lower, this suggests that BN helps the statistical distribution of  deep neural network input values to be more stable . The figure below shows the training and test errors for each round of BN and no BN for the FNFM model:

\begin{figure}[H]
\begin{center}
\includegraphics[scale=0.6]{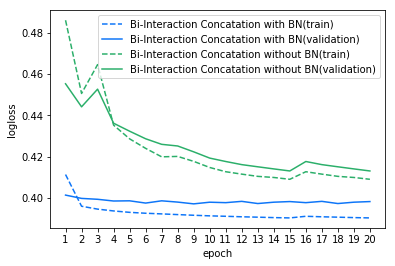}
\end{center}
\vspace*{-10pt}
   \caption{Comparison of FNFM learning curves using BN and not using BN}
\label{fig:why bn}
\vspace*{-12pt}
\end{figure}

Experiments have shown that the training error after using BN is significantly faster than the speed without using BN.
The model using BN can achieve lower cross entropy loss when training same epochs.

\subsection{Model Comparision }

This section compares the FNFM model with the LR, FM, FFM, PNN, WDL, DeepFM, NFM, and DCN models, which include some of the most advanced models currently in the recommended system. In this experiment, because the model is concerned with the automatic learning ability of the feature combination, the features generated by the artificial feature engineering are not added, and all the original features are used.

FNFM combines the FFM and DNN models into an end-to-end model. FFM and DNN use different method when learning high-order combinations of features, one can learn cross feature explicitly by second-order feature combinations, and the other can  learn high-order combinations between features implicitly. FNFM concate the second-order combination features learned by FFM as input to the DNN module which makes it easier for the DNN module to learn the high-order combination patterns contained in the data.

The hyperparameters for each model are obtained by performing a grid search on the validation set. The best parameter settings are given in the corresponding subsections below. For FM and FFM we use the AdaGrad algorithm with an initial learning rate of 0.1, and for other models we optimize with the Adam algorithm with an initial learning rate of 0.0001. Use L2 regularization of size 0.00001 for hidden vectors and embedded vectors. For the FFM and FNFM models, because their parameters are too large, the hidden vectors are fixed in this experiment as 4 dimensions. For other models, search from 4, 8, 16, 32, 64. The structure of the deep neural network is searched from a combination of 2 or 3 layers and 128 or 256 neurons per layer. The following table shows the corresponding test set scores for different models when the validation set achieves the lowest logloss.

\begin{table}[H]
\caption{Results on the Avazu dataset}
\begin{center}
\begin{tabular}{lll}
\hline\noalign{\smallskip}
Model & LogLoss & AUC \\
\noalign{\smallskip}\hline\noalign{\smallskip}
%LR & 0.6377 & -3.36\% \\
LR & 0.4059 & 0.7296 \\
FM & 0.3991 & 0.7433 \\
FFM & 0.3980 & 0.7455 \\
PNN & 0.3983 & 0.7450 \\
WDL & 0.3993 & 0.7425 \\
DeepFM & 0.3981 & 0.7451 \\
NFM & 0.3988 & 0.7437 \\
DCN & 0.3978 & 0.7462 \\
\textbf{FNFM} & \textbf{0.3973} & \textbf{0.7470} \\
\noalign{\smallskip}\hline
\end{tabular}
\end{center}
\end{table}

By observing the above table, it can be seen that LR is the worst effect in the model, which shows that a series of methods based on factorization is useful for modeling sparse features. Since the artificial feature engineering is not performed in this experiment, the WDL model does not perform well, but it has many performance improvements compared to the LR model, which indicates that the deep neural network plays a role in learning high-order feature combinations. The FNFM model is optimal for both the test set Logloss and AUC metrics, which suggests that it is useful to capture high-order features by introducing the concept of the field during feature interaction (comparing with DCN, the concept of fields can reduce confusion during feature interactions) and using the Bi-Interaction Concatation layer (comparing with DFM, it same most information in feature interaction) before the DNN network.

\section{Conclusion}

This paper mainly introduces how to use the deep learning technology to apply to the click-rate prediction task to improve the accuracy of the click-through rate prediction model. From the perspective of feature combination learning, this paper proposes a field-aware neural factorization machine for click rate prediction. This model can have strong second-order feature interactive learning ability like FFM. Further more, deep neural network is used to learn higher-ordered feature combinations. In the experiment on CTR prediction, we first explained that the Bi-Interaction Concatation layer used in the FNFM model has better expressiveness than the Bi-Interaction Pooling layer in the NFM model, and then illustrates the problems and solution in using the Concatation layer. Finally we gave a comparison of the effects of FNFM and other models, which verified its expressive ability, in which we get some hints on the effects by involing factoritation mechanism and information concatations in feature extraction.

Now the prediction model doesn't consider some practical issue in business, including the history of user clicking, change of interests, and the attention which can activate the memory. Our next work is to combine such techniques to make a more useful prediction system.

%% The file named.bst is a bibliography style file for BibTeX 0.99c and we model it as a point-wise ranking problem
\bibliographystyle{named}
\bibliography{fnfm}

\begin{thebibliography}{}

\bibitem[\protect\citeauthoryear{Broder}{2008}]{broder2008computational}
Andrei~Z Broder.
\newblock Computational advertising.
\newblock In {\em SODA}, volume~8, pages 992--992, 2008.

\bibitem[\protect\citeauthoryear{Cheng \bgroup \em et al.\egroup
  }{2016}]{cheng2016wide}
Heng-Tze Cheng, Levent Koc, Jeremiah Harmsen, Tal Shaked, Tushar Chandra,
  Hrishi Aradhye, Glen Anderson, Greg Corrado, Wei Chai, Mustafa Ispir, et~al.
\newblock Wide \& deep learning for recommender systems.
\newblock In {\em Proceedings of the 1st Workshop on Deep Learning for
  Recommender Systems}, pages 7--10. ACM, 2016.

\bibitem[\protect\citeauthoryear{Guo \bgroup \em et al.\egroup
  }{2017}]{guo2017deepfm}
Huifeng Guo, Ruiming Tang, Yunming Ye, Zhenguo Li, and Xiuqiang He.
\newblock Deepfm: a factorization-machine based neural network for ctr
  prediction.
\newblock {\em arXiv preprint arXiv:1703.04247}, 2017.

\bibitem[\protect\citeauthoryear{He and Chua}{2017}]{he2017neural}
Xiangnan He and Tat-Seng Chua.
\newblock Neural factorization machines for sparse predictive analytics.
\newblock In {\em Proceedings of the 40th International ACM SIGIR conference on
  Research and Development in Information Retrieval}, pages 355--364. ACM,
  2017.

\bibitem[\protect\citeauthoryear{Ioffe and Szegedy}{2015}]{ioffe2015batch}
Sergey Ioffe and Christian Szegedy.
\newblock Batch normalization: Accelerating deep network training by reducing
  internal covariate shift.
\newblock {\em arXiv preprint arXiv:1502.03167}, 2015.

\bibitem[\protect\citeauthoryear{Juan \bgroup \em et al.\egroup
  }{2016}]{juan2016field}
Yuchin Juan, Yong Zhuang, Wei-Sheng Chin, and Chih-Jen Lin.
\newblock Field-aware factorization machines for ctr prediction.
\newblock In {\em Proceedings of the 10th ACM Conference on Recommender
  Systems}, pages 43--50. ACM, 2016.

\bibitem[\protect\citeauthoryear{Lee \bgroup \em et al.\egroup
  }{2013}]{lee2013local}
Joonseok Lee, Seungyeon Kim, Guy Lebanon, and Yoram Singer.
\newblock Local low-rank matrix approximation.
\newblock In {\em International Conference on Machine Learning}, pages 82--90,
  2013.

\bibitem[\protect\citeauthoryear{Pepelyshev \bgroup \em et al.\egroup
  }{2015}]{pepelyshev2015adaptive}
Andrey Pepelyshev, Yuri Staroselskiy, and Anatoly Zhigljavsky.
\newblock Adaptive targeting for online advertisement.
\newblock In {\em International Workshop on Machine Learning, Optimization and
  Big Data}, pages 240--251. Springer, 2015.

\bibitem[\protect\citeauthoryear{Rendle}{2010}]{rendle2010factorization}
Steffen Rendle.
\newblock Factorization machines.
\newblock In {\em Data Mining (ICDM), 2010 IEEE 10th International Conference
  on}, pages 995--1000. IEEE, 2010.

\bibitem[\protect\citeauthoryear{Wang \bgroup \em et al.\egroup
  }{2017}]{wang2017adkdd}
Ruoxi Wang, Bin Fu, Gang Fu, and Mingliang Wang.
\newblock Deep \& cross network for ad click predictions.
\newblock In {\em Proceedings of the ADKDD'17}, 2017.

\end{thebibliography}

\end{document}